# MeLIAD: Interpretable Few-Shot Anomaly Detection with Metric Learning and Entropy-based Scoring

Eirini Cholopoulou, and Dimitris K. Iakovidis, *Senior Member, IEEE*

*Abstract*— Anomaly detection (AD) plays a pivotal role in multimedia applications for detecting defective products and automating quality inspection. Deep learning (DL) models typically require large-scale annotated data, which are often highly imbalanced since anomalies are usually scarce. The black box nature of these models prohibits them from being trusted by users. To address these challenges, we propose MeLIAD, a novel methodology for interpretable anomaly detection, which unlike the previous methods is based on metric learning and achieves interpretability by design without relying on any prior distribution assumptions of true anomalies. MeLIAD requires only a few samples of anomalies for training, without employing any augmentation techniques, and is inherently interpretable, providing visualizations that offer insights into why an image is identified as anomalous. This is achieved by introducing a novel trainable entropy-based scoring component for the identification and localization of anomalous instances, and a novel loss function that jointly optimizes the anomaly scoring component with a metric learning objective. Experiments on five public benchmark datasets, including quantitative and qualitative evaluation of interpretability, demonstrate that MeLIAD achieves improved anomaly detection and localization performance compared to state-of-the-art methods.

## I. Introduction

ANOMALY detection (AD) refers to the process of recognizing patterns in data that deviate from normal, a process that can be of high practical significance in diverse multimedia applications, including video surveillance and defect detection [1]. AD is considered a key component in multimedia-based quality inspection that helps assure high production quality and cost efficiency in identifying and discarding defective products. Also, it is particularly challenging considering that in real-world applications, anomalous instances are scarce, heterogenous, and not readily available compared to normal data.

Considering the data availability and heterogeneity of anomalies, various AD methods have been developed requiring different levels of supervision. Fully supervised methods assume all possible defect classes to be known beforehand; therefore, they are unsuitable in real-world scenarios where not all defect classes are known a priori. To address this limitation supervised one-class classification AD methods have been proposed, capable of learning from normal samples to discriminative samples that deviate from normal [2]. Also, various unsupervised AD methods have been proposed [3], [4]. However, these approaches often incur a high false positive rate [5]. Few-shot learning has been proposed as a less demanding supervised approach for AD, capable of learning to discriminate anomalies from only a few annotated training samples that do not necessarily belong to one class [6], [7] . Interestingly, it has been recently shown that some of the few shot AD methods are capable of generalizing well, even to new, previously unknown, anomaly classes [8]. However, such methods often impose prior distribution assumptions that may deviate from that of the true anomalies, limiting their generalization capacity to unseen classes. Although this is a promising direction towards real-world AD, the works tackling this challenge are still limited.

Furthermore, in AD it is often required to have some cues explaining why a sample is characterized as anomalous. This means that the model should not only identify anomalies but also provide visual explanations that highlight the specific features or regions of the data that contributed to the anomaly detection. This can be particularly useful in AD systems, *e.g.*, to adhere to fairness requirements or to investigate possible bias effects [9]. Methods with such a capacity are characterized as interpretable. In the context of AD only a limited number of such methods have been proposed [10], [11]. However, most current approaches address interpretability not inherently, but using simplified post-hoc models. These models do not consider the computations of the original model, and their use may lead to unreliable interpretations [12].

In this study, we propose a novel Metric Learning and Entropy-based methodology for Interpretable Anomaly Detection (MeLIAD) in images. This methodology is motivated by the need of trustworthy methods, inherently able to reveal the rationale behind the anomaly detection and scoring, without any prior assumptions on the distribution of the true anomalies. It is based on Convolutional Neural Networks (CNNs) and it requires only a few samples from a subset of anomaly classes for training. MeLIAD derives anomaly scores by using an entropy-based measure to identify the most informative feature maps, by assessing their activation probability to identify the level of abnormality in images. The regions in the images with the highest entropy scores correspond to the most anomalous regions, which are highlighted in the form of interpretation

Eirini Cholopoulou, and Dimitris K. Iakovidis are with the University of Thessaly, Department of Computer Science and Biomedical Informatics, Lamia 35131 Greece (e-mail: {echolopoulou, diakovidis}@uth.gr).

heatmaps. The proposed anomaly scoring component is optimized during training to assign higher entropy values to images with anomalies than to normal images. This renders MeLIAD inherently interpretable by enabling an enhanced representation of the anomalies within the CNN feature maps, which is used to visually explain why some image regions are characterized as anomalous. The contributions of this work can be summarized as follows:

- To the best of our knowledge, MeLIAD constitutes the first inherently interpretable metric learning based methodology, that does not impose any prior assumption of true anomalies, for anomaly detection applications.
- A novel anomaly scoring component, which unlike current components of this kind, it is entropy-based and trainable to identify and highlight the most informative image regions, considered as anomalous, contributing to the inherent interpretability of MeLIAD.
- A novel loss function is introduced that combines an adaptive margin-based loss, ensuring accurate performance, with the anomaly scoring loss that drives interpretability in an end-to-end manner.
- A comprehensive experimental study is performed, that includes the quantitative and qualitative evaluation of interpretability, comparing various state-of-the-art few-shot learning and interpretable AD methods, highlighting the advantages of the proposed methodology.

The rest of this manuscript is organized as follows. Section II provides an overview of the related work. The proposed MeLIAD methodology is detailed in section III, and the results from its comparative experimental evaluation are presented in section IV. Lastly, section V summarizes the conclusions derived from this study.

## II. RELATED WORK

Traditional AD methods usually leverage statistical measures and information theory [13]. These methods can be effective; however, they struggle to adapt in complex and high-dimensional data representations. More recently, deep learning methods with automatic feature extraction capabilities, have been proposed to offer enhanced adaptivity in solving AD problems [14]. Some of these approaches usually leverage pre-trained deep learning models only for the task of feature extraction, and handle the anomaly scoring process as a separate task [15]. Recently, the joint optimization of feature extraction with anomaly scoring has been proved a promising direction providing improved AD performance; however, there is still only a limited work, with the current methods being fully supervised [16].

### A. Few-shot Anomaly Detection

Few-Shot AD (FSAD) methods are becoming increasingly popular in anomaly detection multimedia applications due to their fewer training requirements [17]. Some FSAD approaches leverage energy-based generative models to synthesize defective samples based on the available ones [18], and hierarchical generative models (HGM) for this purpose [19]. A recent approach, called FastRecon [20], utilizes a few normal samples to reconstruct the normal versions of the anomalies, and then AD is achieved by sample alignment. There is a limited number of FSAD methods that do not require all classes to be represented in the training set. Among these, Registration-based few-shot Anomaly Detection (RegAD) [21] employs a registration-based scheme to train a category-agnostic learning model, and Deviation Networks (DevNet) [22] is based on deviation learning that leverages prior probabilities. These methods have demonstrated a remarkable performance outperforming the most recent approaches.

### B. Interpretable Anomaly Detection

In AD interpretability refers to the extraction of useful insights from a model into why specific data are identified as anomalous [23]. The demand for interpretation of decisions made by black box systems, such as artificial neural networks, has led to the use of model-agnostic tools that offer insights about deep learning-based AD in a post-hoc manner *i.e.*, applied after the training process is complete [24]. Examples of interpretable AD (IAD) methods using post hoc techniques include attention-based models [25] and reconstruction-based methodologies [11], that provide saliency maps for individual predictions.

Methodologies that reveal the rationale behind the anomaly scoring mechanism, thus incorporating interpretability directly into the model architecture, are characterized as inherently interpretable or interpretable by design [12]. In AD, such approaches can ensure interpretations that are more coherent with how the anomaly scores drive the decisions of the machine learning models [26]. Inherently IAD approaches include Fully Convolutional Data Description (FCDD) [27], which employs a gradient-based mechanism to generate anomaly heatmaps. Similarly, multiresolution knowledge distillation (MKD) [28] utilizes a cloner network to distill feature information and produce anomaly interpretation heatmaps. DevNet [22], learns prior-driven anomaly scores and provides interpretations by attributing them to the inputs through gradient backpropagation.

MeLIAD is inspired by the most recent methods targeting real-world applications, such as DevNet, which do not require training samples from all possible anomaly classes. Most AD methods, incorporating anomaly scoring components, e.g., [29], are based on specific distance measures to identify anomalies, and they are not considered inherently interpretable, since the anomaly scores are not directly optimized by the model. Unlike the previous methods, MeLIAD introduces an entropy-based anomaly scoring mechanism, that is trained jointly with metric learning to optimize the entire AD process in an end-to-end inherently interpretable manner. This scoring component is used to generate visual interpretations that indicate why an image was identified as anomalous, providing insights into the decision-making process of the model.

## III. METHODOLOGY

### A. Overview

An overview of MeLIAD is schematically illustrated in Fig. 1. It is a few-shot learning methodology that consists of three components, namely the feature extraction, the anomaly scoring, and the anomaly interpretation component. During its training phase (dotted lines), the feature extraction is optimized by a metric learning objective, the anomaly scoring component is optimized by a novel entropy-based loss function, and both components are adapted using another loss function designed to jointly optimizing the AD performance in an end-to-end manner.

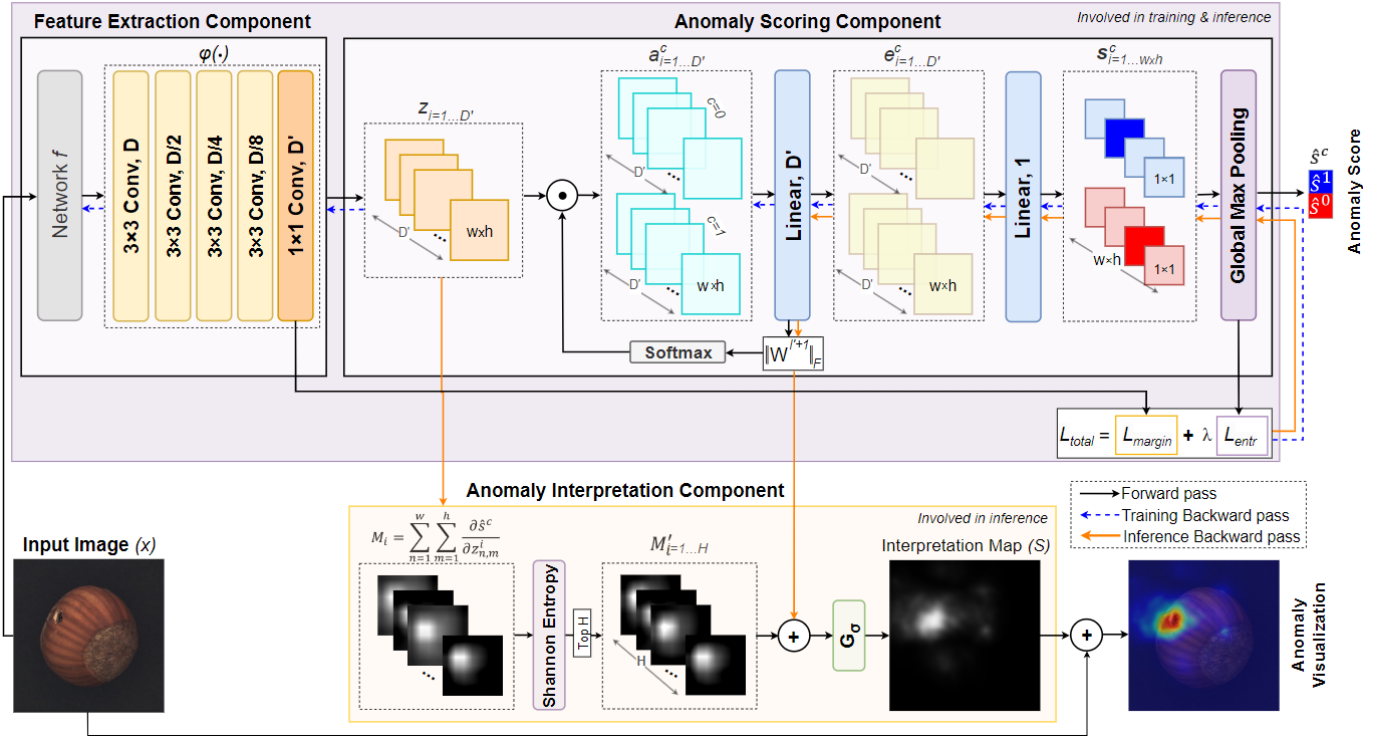

Fig. 1 Overview of the MeLIAD methodology.

The first component implementing feature extraction includes a pre-trained network $f$ and a feature reduction block $\varphi$. Network $f$ receives an image $x$ as input and outputs a set of $D$ feature maps with size $w \times h$ from its last convolutional layer. Block $\varphi$ aims to reduce the computational load of the next processing steps by reducing the number of feature maps to $D' < D$ with the application of a series of convolutions, while maintaining their size.

The second component $\eta$, implements anomaly scoring, *i.e.*, it assigns a score quantifying the degree to which the input image can be characterized as anomalous. This is performed in two stages: a) the feature maps $z_i, i = 1, \ldots, D'$ are transformed to embeddings $e_i^c, i = 1, \ldots, D', c = 0,1$, where $c$ represents the class of the input image (0 normal / 1 abnormal). These embeddings aim to provide representations that are more focused on the anomalies. They are computed by a linear transformation of the feature maps $a_i^c, i = 1, \ldots, D'$, which are class-wise reweighted instances of the feature maps $z_i$, highlighting the features that discriminate their contents upon the class they belong to; b) the embeddings $e_i^c$ are then reduced by applying a linear transformation along the depth of the tensor, to derive a set of anomaly scores $\mathbf{s}^c = (s_1^c, \ldots, s_{w \times h}^c), c = 0,1$. Specifically, the output vectors $\mathbf{s}^c$ combine the information from the $D'$ embeddings into a single output value for each spatial location within the embeddings to indicate the likelihood of an anomaly being present. The final anomaly score $\hat{s}^c$ is derived by applying a global max pooling operation on the derived anomaly scores, so that the highest anomaly score is selected. If the highest anomaly score $\hat{s}^c$ corresponds to $c = 1$ the image is classified as anomalous, otherwise its classified as normal.

The anomaly interpretation component generates a map $S$, by backpropagating the gradients of the loss (continuous orange arrows) through the network with respect to the input image $x$. This interpretation heatmap highlights the most active regions contributing to the prediction, by selecting and aggregating the top $H$ activation maps with the highest Shannon entropy scores.

During inference, MeLIAD receives a test image $x$ as input and maps it to the feature representation space, using the distance learned from the metric learning process. Then the anomaly scoring component assigns an anomaly score to the test image based on its learned weights. In the case of $\hat{s}^1 > \hat{s}^0$ the test image is classified as an anomaly. Finally, the anomaly interpretation component of MeLIAD highlights parts of the test image that are identified as possibly anomalous. The synergic integration of all components not only addresses the challenges of few-shot AD, but also fulfills the demand for interpretability in quality inspection scenarios. The individual components of the proposed methodology are further detailed in the following sections.

### B. Problem Statement

This study addresses the challenge of detecting anomalies by leveraging a limited set of labelled images, that includes only a subset of a few known anomalous instances. Let us consider a training set $X = \{x_1, x_2, \ldots, x_N, x_{N+1}, \ldots, x_{N+A}\}$, where the majority of the labeled images are normal, $X_n = \{x_1, x_2, \ldots, x_N\}$, and only a limited set of them are abnormal, $X_a = \{x_{N+1}, x_{N+2}, \ldots, x_{N+A}\}$, $A \ll N$. The objective is to learn a function $\tau(\cdot): X \to \mathbb{R}$ that assigns an anomaly score $\hat{s}^c = \tau(x)$ to a sample $x$, where $c$ represents the class (0 normal / 1 abnormal), such that $\hat{s}^1 > \hat{s}^0$.

## C. Feature Extraction and Metric Learning

Feature extraction can be implemented by any given neural network $f$ that learns to map input images $(x)$ to lower-dimensional representations $f(x)$. The network is followed by a feature reduction block $\varphi(\cdot; \Theta_r)$, with $\Theta_r = \{W^l | l = 1, \ldots, l'\}$ represent its trainable parameters, where $W^l$ is the weight matrix corresponding to the convolutional layer $l$. This block consists of four convolutional layers using Rectified Linear Units (ReLU) activation functions, each followed by batch normalization. The final layer of the reduction block is a 1×1 convolutional layer that reduces the dimensionality to the desired $D'$ output dimension.

Generally, in deep metric learning the objective is to learn a highly separable embedding space, in which the distance between feature representations of the same class is minimized and maximized for representations of different classes. Given two feature maps $z_i = \varphi(f(x_i))$ and $z_j = \varphi(f(x_j))$ the distance metric in the embedding space can be calculated as:

$$d_{ij} = \|z_i - z_j\| \qquad (1)$$

where $\|\cdot\|$ represents the cosine distance.

In order to optimize the metric learning objective to effectively separate normal data from anomalies, the adaptive margin-based loss [30] is adopted to facilitate the learning process of the distance metric defined in (1). This loss was chosen because it is well-recognized for its effectiveness in learning pairwise relationships, by leveraging distance weighted sampling to select more informative datapoints than traditional approaches. It is defined as:

$$l_{margin} := max(0, (\mu + y_{ij}(d_{ij} - \beta)) \qquad (2)$$

where $y_{ij} \in \{0,1\}$ indicates if a pair of samples is positive ($y_{ij} = 1$), or negative ($y_{ij} = 0$), $\mu$ is the hyperparameter that defines the minimum margin for the separation between positive and negative pairs and $\beta$ determines a shift of the margin boundary. During training, the minimization of (2) encourages the network to adjust its trainable parameters to learn the feature representations $z_i$ by assigning smaller distance values between positive pairs and larger distance values between negative pairs by a margin, ultimately enhancing the capacity of the model to identify anomalies.

## D. Anomaly Scoring

The purpose of the anomaly scoring component is to facilitate inherently interpretable anomaly learning by encouraging the model to identify anomalous feature representations and derive visual interpretations. The anomaly scoring process in the proposed methodology is defined as a mapping function $\eta(\cdot): \mathbb{R}^{D'} \to \mathbb{R}$ that is performed in two stages. The first one includes the transformation of the feature representations $z_i$, $i = 1, \ldots, D'$ to embeddings $e_i^c$, $i = 1, \ldots, D', c = 0,1$. These embeddings can be formulated as a linear transformation:

$$e_i^c = W^{l'+1} a_i^c + \boldsymbol{b}^c \qquad (3)$$

with trainable parameters $\Theta_u = \{W^{l'+1}, \boldsymbol{b}^c\}$, where $W^{l'+1} \in \mathbb{R}^{D' \times D'}$ is a weight matrix, and $\boldsymbol{b}^c = (b_1^c, \ldots, b_{D'}^c)$ is a vector composed of bias terms. The embeddings $a_i^c$, $i = 1, \ldots, D'$ are computed as the elementwise product between the feature representations $z_i$, $i = 1, \ldots, D'$ and the probability distribution of the weights $W^{l'+1}$ that indicate the relevance of specific patterns to each respective class:

$$a_i^c = z_i \odot \sigma\left(\frac{\left\|W^{l'+1}\right\|_F}{t}\right) \qquad (4)$$

where $\sigma$ represents the SoftMax operation, $\odot$ denotes the elementwise product, $\|\cdot\|_F$ denotes the Frobenius norm, and $t$ is a regularization temperature parameter. The output of the SoftMax is expanded along the spatial dimensions to $\mathbb{R}^{w \times h \times D'}$ to match the dimensionality of $z_i$, in order to perform the element-wise operation. The use of the SoftMax activation function aims to highlight the most relevant patterns to a class, by amplifying the higher weight values and diminishing the lower values. Thus, the values of the reweighted feature maps $a_i^c$ signify the abnormality degree of each feature map, with higher values ($t \to 0$) indicating increased involvement of anomalous patterns in an image.

The next step involves the computation the anomaly scores $\boldsymbol{s}^c$, by reducing the embeddings $e_i^c$ along the depth dimension, to compress their information and focus on the most relevant features for anomaly detection. Each element in $\boldsymbol{s}^c = (s_1^c, \ldots, s_{w \times h}^c)$, is obtained by a linear combination of the corresponding elements along the depth of the tensor $e_i^c$, expressed as:

$$\boldsymbol{s}^c = W^{l'+2} e_i^c + b \qquad (5)$$

with trainable parameters $\Theta_s = \{W^{l'+2}, b\}$, where $W^{l'+2} \in \mathbb{R}^{1 \times D'}$ is the weight matrix and $b \in \mathbb{R}$ the bias parameter. Then the final anomaly score $\hat{s}^c$ is derived by selecting the maximum value across the depth of the anomaly scores tensor.

Therefore, the overall process of mapping an anomaly score to an input image $x$ can be addressed in an end-to-end manner, expressed as an anomaly score learning function $\tau(\cdot; \Theta): X \to \mathbb{R}$, that can be represented as:

$$\tau(\cdot; \Theta) = \eta((\varphi; \Theta_r); \Theta_u, \Theta_s), \qquad (6)$$

where $\Theta = \{\Theta_r, \Theta_u, \Theta_s\}$ are the respective trainable parameters of MeLIAD.

## E. Loss Function for Joint Optimization

A loss function is proposed to facilitate the construction of a feature representation space by learning the distance metric defined in (1), while simultaneously refine the anomaly scores to align with the objective of the anomaly scoring function $\tau$. This loss function aims to enhance the robustness of the model in tasks characterized by imbalanced class distributions.

To this end, the hypersphere classification (HSC) loss [31], is adapted so that it leverages the entropy to optimize the derived anomaly scores, instead of optimizing the Euclidean distance of the mapped feature representations as anomaly scores. Given the anomaly scores $\hat{s}^c$ (predicting class $c$) and target labels $y \in \{0,1\}$, where $y = 1$ denotes an anomalous sample and $y = 0$ denotes a normal sample, the loss function can be expressed as:

$$l_{entr} = \frac{1}{N+A}\sum_{i=1}^{N+A}(1-y_i)\hat{s}^c - y_i \log(1-e^{-\hat{s}^c}) \qquad (7)$$

This loss function penalizes higher anomaly scores for normal samples ($y_i = 0$) and low anomaly scores for anomalous samples ($y_i = 1$), which is consistent with the condition $\hat{s}^1 > \hat{s}^0$ of the anomaly score learning function $\tau$.

The joint optimization of both objectives can be achieved in an end-to-end manner, by the minimization of the following loss function formulation:

$$L_{total} = l_{entr} + \lambda l_{margin} \qquad (8)$$

where $\lambda$ is a positive hyperparameter that controls the impact of the $l_{margin}$ objective. The first component of the loss function, $l_{entr}$, refines the anomaly scores to receive higher values for images with anomalies, and lower values for normal images, by minimizing the discrepancy between the predicted and the true anomaly scores. The second component, $l_{margin}$, learns a distance metric to effectively map the feature representations of the input data. The combined effect of these two losses contributes to the holistic optimization of the model. Higher values of lambda place more emphasis on the margin loss and encourage the model to learn decision boundaries that are more discriminative, whereas lower values of lambda place more emphasis on the $l_{entr}$ loss. Thus, $l_{entr}$ encourages the model to use the most informative feature representations towards assigning anomaly scores that predict the degree to which data samples deviate from normal.

### F. Anomaly Interpretation

The inherent interpretability of the model, involves using the anomaly scoring mechanism to derive interpretations, by selecting the most informative feature maps of an image. The regions in the feature maps with the highest entropy-based anomaly scores, correspond to the most anomalous regions, which are highlighted in the form of interpretation heatmaps. The generation of an interpretation heatmap $S$ highlights the most active regions of the input image that contribute to its classification as anomalous. More specifically, the anomaly interpretation process of MeLIAD involves the backward propagation of the loss through the network with respect to the feature map activations $z_i$, to receive $M_i, i = 1, \dots, D'$ activation maps which are formulated as:

$$M_i = \sum_{n=1}^{w}\sum_{m=1}^{h}\frac{\partial \hat{s}^c}{\partial z^i_{n,m}} \qquad (9)$$

where $z^i_{nm}$ is the $i^{th}$ feature map with indices $n = 1, \dots, w$ and $m = 1, \dots, h$ denoting the spatial positions of width and height respectively and $\hat{s}^c$ refers to the score for the class $c$, with the gradients of the abnormal class (target class) set to 1 and for the normal class set to zero.

TABLE I: ANOMALY DETECTION DATASETS USED IN THIS STUDY.

| Dataset | Domain | Classes | Anomaly Types | Images | Resolution |
|---|---|---|---|---|---|
| *MVTec AD* [32] | Industry/ Defect Inspection | 15 | 73 | 5,354 | 700×1024 |
| *KolektorSDD*[35] | Industry/ Surface Defect | 1 | 5 | 399 | 500×1270 |
| *GoodsAD* [33] | Supermarket Goods / Defect detection | 6 | 15 | 6,124 | 3000×3000 |
| *Mastcam* [36] | Multispectral / Novelty Detection | 1 | 8 | 9,728 | 64×64 |
| *MSD* [34] | Mobile Phone screens / Defect Detection | 1 | 3 | 1.200 | 1080×1920 |

The interpretation heatmap is then derived as $S = G_\sigma(\sum_{i=1}^{H} M'_i)$ by aggregating the most informative activation maps $M'_i$. These maps correspond to the top $H$ entropy scores of the respective activations $M_i$, computed by the Shannon entropy measure. $G_\sigma(\cdot)$ denotes a 2D Gaussian filter with standard deviation $\sigma$, that is utilized for noise reduction and smoothing. To visualize the interpretation heatmap, $S$ is upsampled with bilinear interpolation to match the size of the original input image $x$.

## IV. EXPERIMENTS AND RESULTS

### A. Experimental Setup

The proposed methodology was thoroughly evaluated using 5 public AD datasets presented in Table I. The MVTec AD [32] dataset, is a large publicly available benchmark dataset widely acknowledged as a reference for industrial anomaly detection. It consists of 5,354 images of 15 substantially different object and texture classes, comprising a set of 3,629 normal images and a set of 1,725 images with various kinds of anomalies, including bottles, cables, capsules, hazelnuts, metalnuts, pills, screws, toothbrushes, transistors, zippers, and images of carpet, grid, leather, tile and wood textures. Each category is divided into a training set, that contains only normal images, and a test set that contains both normal images and images of various types of anomalies. The images vary in resolution, ranging from 700×700 to 1024×1024 pixels. The recent GoodsAD dataset [33], consists of six categories of supermarket goods, including boxed cigarettes, bottled drinks, canned drinks, bottled foods, boxed foods, and packaged foods. Each category has several types of anomalies. GoodsAD includes 6,124 images, with 4,464 images of normal goods and 1,660 images of anomalous goods. The Mobile phone screen Surface Defect (MSD) dataset [34], consists of 1,200 images that depict three types of defects: oil, stain and scratch. The Kolektor Surface-Defect Dataset (KolektorSDD) [35] consists of 399 images from which 52 have surface defects and 347 images are normal. Mastcam [36] is a dataset of 9,728 hyperspectral images of geologic observations on Mars, from which 426 images were classified into 8 types of novel classes that include meteorite, float rock, bedrock, vein, broken rock, dump pile, drill hole, and dust removal tool. All datasets were used as provided from their sources, without

TABLE II: Comparative Anomaly Detection Results of State-of-the-Art FSAD Methods for Different Numbers of Anomalous Shots.

| k-shot | Methodologies | | | | | | | | | | | |
|---|---|---|---|---|---|---|---|---|---|---|---|---|
| | Image-level AUROC | | | | | | Pixel-level AUROC | | | | | |
| | HGM | DevNet | PatchCore | RegAD | FastRecon | MeLIAD | HGM | DevNet | PatchCore | RegAD | FastRecon | MeLIAD |
| $k=2$ | 0.751 | 0.836 | 0.881 | 0.857 | **0.901** | 0.889 | N/A | 0.847 | 0.929 | 0.941 | **0.949** | 0.941 |
| $k=4$ | 0.761 | 0.861 | 0.898 | 0.882 | 0.928 | **0.936** | N/A | 0.856 | 0.939 | 0.955 | 0.957 | **0.960** |
| $k=8$ | 0.783 | 0.911 | 0.923 | 0.912 | 0.941 | **0.956** | N/A | 0.886 | 0.943 | 0.963 | 0.962 | **0.964** |

*The best results are highlighted in bold.

TABLE III: Comparative Anomaly Detection Results per Image Category of State-of-the-Art FSAD and IAD Methods in Terms of Image-Level (*IL*) and Pixel-Level (*PL*) AUROC Scores.

| Category | Methodologies | | | | | | | | | | | | | | | | | | |
|---|---|---|---|---|---|---|---|---|---|---|---|---|---|---|---|---|---|---|---|
| | FSAD | | | | | | | | | | | IAD | | | | | FSAD+IAD | | |
| | HGM | | PatchCore | | FSFA | | RegAD | | FastRecon | | | FCDD | | MKD | | DevNet | | MeLIAD | |
| | IL | PL | IL | PL | IL | PL | IL | PL | IL | PL | | IL | PL | IL | PL | IL | PL | IL | PL |
| Carpet | 0.642 | N/A | 0.988 | **0.992** | N/A | 0.840 | 0.985 | 0.983 | **0.999** | 0.991 | | 0.899 | 0.962 | 0.748 | 0.942 | 0.953 | 0.921 | 0.989 | 0.975 |
| Grid | 0.798 | N/A | 0.737 | 0.823 | N/A | 0.820 | 0.915 | 0.866 | 0.925 | 0.873 | | 0.708 | 0.903 | 0.781 | **0.909** | 0.928 | 0.889 | **0.962** | 0.895 |
| Leather | 0.979 | N/A | **1.000** | 0.950 | N/A | 0.950 | **1.000** | 0.993 | **1.000** | 0.991 | | **1.000** | 0.982 | 0.929 | 0.959 | 0.956 | 0.973 | 0.889 | 0.983 |
| Tile | 0.816 | N/A | **1.000** | 0.972 | N/A | 0.760 | 0.996 | 0.961 | 0.998 | 0.948 | | 0.978 | 0.929 | 0.881 | 0.796 | 0.945 | 0.903 | 0.985 | **0.973** |
| Wood | 0.961 | N/A | 0.936 | 0.861 | N/A | 0.780 | **0.995** | 0.964 | 0.989 | 0.953 | | 0.986 | 0.897 | 0.972 | 0.836 | 1.000 | 0.854 | **0.995** | 0.962 |
| Bottle | 0.905 | N/A | 0.997 | **0.991** | N/A | 0.820 | 0.997 | 0.975 | 0.998 | 0.984 | | 0.990 | 0.971 | 0.964 | 0.892 | 0.967 | 0.915 | **0.999** | 0.972 |
| Cable | 0.766 | N/A | **0.969** | 0.982 | N/A | 0.800 | 0.815 | 0.948 | 0.938 | **0.976** | | 0.779 | 0.917 | 0.828 | 0.922 | 0.866 | 0.933 | 0.892 | 0.969 |
| Capsule | 0.618 | N/A | 0.831 | 0.954 | N/A | 0.900 | 0.784 | 0.962 | 0.895 | 0.982 | | 0.658 | 0.930 | 0.787 | 0.87 | 0.759 | 0.883 | **0.948** | **0.991** |
| Hazelnut | 0.811 | N/A | 0.930 | 0.978 | N/A | 0.940 | 0.973 | 0.984 | 0.942 | 0.972 | | 0.951 | 0.957 | 0.943 | 0.943 | 0.999 | 0.924 | **0.979** | **0.989** |
| Metalnut | 0.753 | N/A | 0.911 | 0.966 | N/A | 0.780 | 0.985 | **0.983** | 0.946 | 0.968 | | 0.916 | 0.851 | 0.714 | 0.869 | 0.971 | 0.828 | **0.951** | 0.964 |
| Pill | 0.666 | N/A | 0.904 | 0.899 | N/A | 0.880 | 0.778 | **0.967** | 0.894 | 0.972 | | 0.775 | 0.795 | 0.814 | 0.86 | 0.785 | 0.776 | **0.906** | 0.958 |
| Screw | 0.75 | N/A | 0.733 | 0.949 | N/A | 0.860 | 0.657 | 0.953 | 0.793 | 0.930 | | 0.689 | 0.887 | 0.792 | 0.947 | 0.912 | 0.798 | **0.939** | **0.971** |
| Toothbrush | 0.677 | N/A | 0.964 | 0.964 | N/A | 0.850 | 0.966 | 0.989 | 0.878 | 0.981 | | 0.744 | 0.927 | 0.914 | 0.941 | 0.863 | 0.885 | **0.991** | 0.989 |
| Transistor | 0.793 | N/A | **0.970** | 0.883 | N/A | 0.800 | 0.902 | 0.949 | 0.959 | **0.951** | | 0.885 | 0.877 | 0.852 | 0.728 | 0.89 | 0.890 | 0.945 | 0.899 |
| Zipper | 0.814 | N/A | 0.977 | 0.977 | N/A | 0.860 | 0.934 | 0.973 | 0.963 | 0.965 | | 0.950 | 0.928 | 0.923 | 0.919 | 0.868 | 0.914 | **0.979** | 0.971 |
| Average | 0.783 | N/A | 0.923 | 0.943 | N/A | 0.843 | 0.912 | 0.963 | 0.941 | 0.962 | | 0.861 | 0.914 | 0.856 | 0.889 | 0.911 | 0.886 | **0.956** | **0.964** |

*The best results are highlighted in bold.

employing any data augmentation technique. Pixel-level ground truth annotations are provided for images with anomalies. In our experiments, for each category the training set of normal images and a very small sample of $k$ randomly selected images with anomalies from the test set (see Section IV.B), were used for training. The rest of the test set was used for testing, i.e. the test set images without the $k$ anomalous samples.

To quantify the AD performance of the proposed methodology the Area Under the Receiver Operating Characteristic (AUROC) was adopted as the most widely used metric in the context of AD. AUROC was used both for the detection of images (at image-level) with anomalies and for anomaly localization (at pixel-level).

To quantitatively assess the interpretability, the well-recognized Remove and Debias (ROAD) metric was used [37]. ROAD measures the consistency, debias and robustness of an attribution method in its capacity to provide heatmap explanations. This is achieved by evaluating the shift in the classification confidence of the model, when modifying the regions of the image indicated by the heatmap. Methods with higher ROAD scores are considered more effective in providing interpretations. The proposed methodology was implemented in Python 3.6 and Pytorch 1.10. Network $f$ was implemented using VGG-11 CNN (without fully connected layers) pre-trained on a generic dataset (ImageNet) – not designed specifically for anomaly detection – from a wide variety of domains and contexts (e.g., animals, objects, scenes). The Adam optimization algorithm was used for training, with an initial learning rate set to 1e-3. The batch size for training was set to 32. The early stopping technique was used for training and the maximum number of epochs was set to 200. All images were resized to 224×224 pixels, considering the expected input size of the VGG-11 network. For the margin-based loss the parameters $\mu = 0.2$ and $\beta = 1.2$ were used as suggested in [30]. A grid search strategy with values in the range of [0, 2] and step of 0.5, was employed for the selection of the hyperparameter $\lambda$, that represents the trade-off term of (8), with the objective to jointly maximize the detection and interpretability of the model. For $\lambda < 1$ and $\lambda > 1$, the AD performance in terms of AUROC was deteriorating by 0.02 to 0.04, as compared with the performance achieved for $\lambda = 1$, which was the best. A 2D Gaussian filter $G_\sigma$ was applied on the derived interpretation heatmap, with a standard deviation $\sigma = 4$.

*B. Comparative Evaluation*

The performance of MeLIAD was compared with several state-of-the-art FSAD methods, including PatchCore [38], RegAD [21], FSFA [18], FastRecon [20] and HGM [19], as well as inherently IAD methods including MKD [28], DevNet [22], and FCDD [27]. The quantitative and qualitative results were obtained by using the official source code and hyperparameters of each reported method.

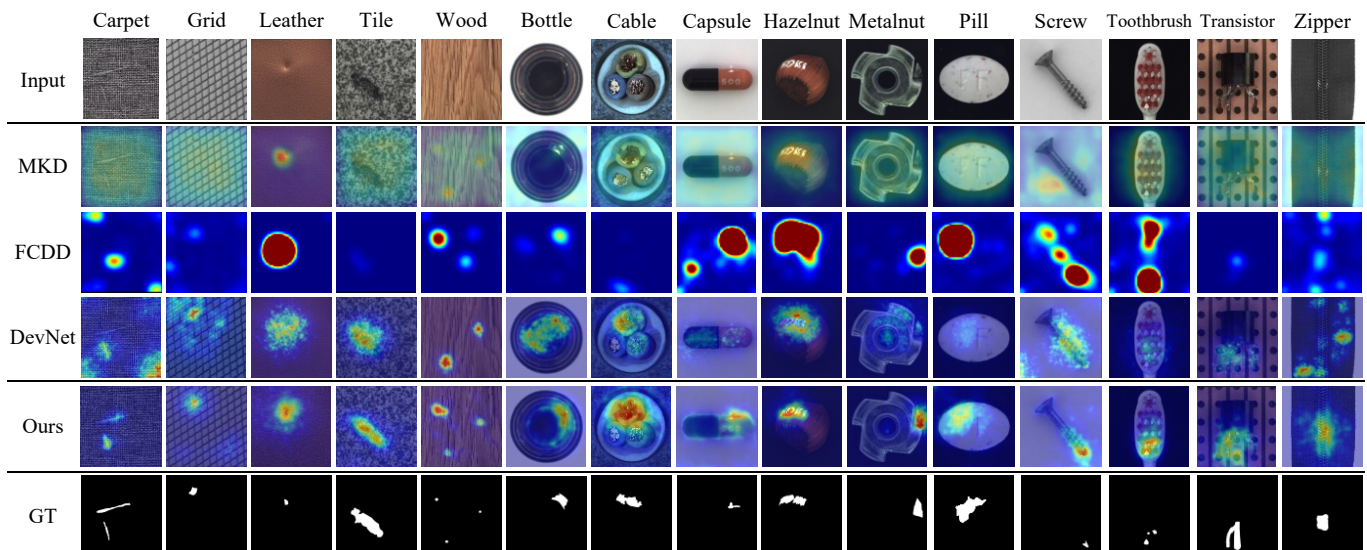

Fig. 2. Interpretation heatmaps for anomaly localization on the MVTec AD dataset. Comparisons are performed among all inherently IAD methods, including MKD, FCDD and DevNet. GT denotes the ground truth masks.

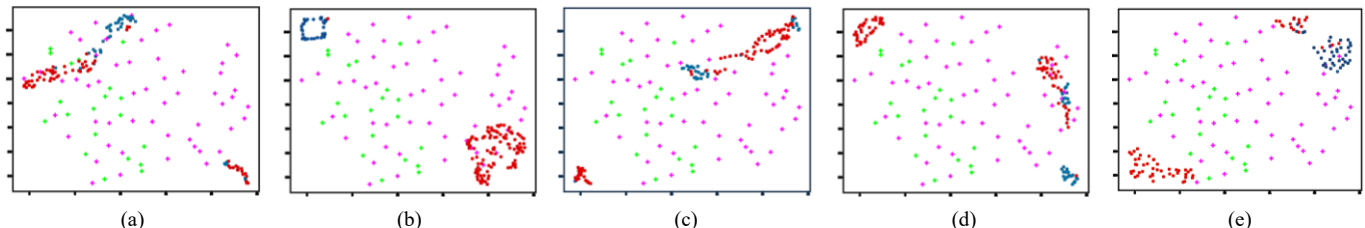

Fig. 3. T-SNE feature representations plots of the test images of the dataset. Each plot includes anomaly (purple, red) and normal (green, blue) data, before (green and purple crosses) and after (blue and red circles) the optimization of the metric learning objective. Indicative classes include: (a) Carpet. (b) Grid. (c) Screw. (d) Tile. (e) Hazelnut.

The experiments were performed separately for each category in the dataset, using all normal images of the training set and a very small subset of $k$ randomly selected images with anomalies. The rest of the anomalous images were used for testing. To determine the value of $k$ an ablation study was conducted. The results are presented in Table II. It can be noticed that MeLIAD achieves the highest average AUROC scores for the 4-shot and 8-shot scenarios, for both detection and localization of anomalies, compared to all the compared FSAD methods. In the 2-shot scenario, MeLIAD demonstrates a lower performance, when compared to the FastRecon methodology; however, the performance of FastRecon heavily depends on the feature reconstruction quality of the normal samples and it does not provide interpretations of its results. The rest of the experiments were performed using $k=8$, because for this setting all the compared methodologies provide an image-level AUROC > 90%.

More detailed results, per image category, were performed to investigate the performance of the proposed methodology in comparison to state-of-the-art FSAD and IAD methods. The results with respect to the detection of images containing anomalies are summarized in Table III. It can be noticed that MeLIAD achieves the highest average AUROC score, which is 95.6%. Specifically, MeLIAD achieved better classification results among all methods in 10 out of 15 classes, outperforming PatchCore, RegAD, FastRecon and HGM by 3.3%, 4.4%, 1.5% and 17.3%, respectively. For the remaining 5 classes the best results were achieved by the PatchCore and FastRecon methods. Furthermore, in comparison to state-of-the-art IAD methodologies, MeLIAD demonstrated better performance among all categories, except for the leather class. MeLIAD outperformed MKD, DevNet and FCDD by 10%, 4.5% and 9.5% respectively. Respective results were obtained comparing the capacity of the proposed methodology to localize anomalies within the images. The results of this evaluation are presented in Table III. It can be noticed that MeLIAD outperformed all the IAD methods, reaching 96.4%. Further analyzing the results contentwise, MeLIAD achieved an average score of 95.7% for the texture classes. This is higher than RegAD and FCDD, which are the best performing FSAD and IAD methods, respectively. For the object classes, the average AUROC score for MeLIAD is 96.7%, outperforming the localization results reported for all FSAD and IAD methods. MeLIAD demonstrates a lower performance for the classes of transistor and grid, that are generally characterized by periodic background patterns and structures.

The average results of the best performing IAD and FSAD methods on all datasets considered in this study are summarized in Table IV. It can be noticed that MeLIAD performs better or comparable (in the case of KolektorSDD) to the other methods across all datasets.

### C. Interpretability Evaluation

The interpretability evaluation of MeLIAD was assessed both qualitatively (Fig. 2), presenting visual interpretation heatmaps, and quantitatively (Table V), reporting the results of the ROAD interpretability quantification metric. The qualitative evaluation involves the comparison with the state-of-the-art inherently

TABLE IV: AVERAGE ANOMALY DETECTION RESULTS IN TERMS OF IMAGE-LEVEL AUROC SCORES ACROSS ALL AD DATASETS AMONG THE BEST IAD AND FSAD METHODS.

| Method | Dataset | | | | |
|---|---|---|---|---|---|
| | MVTec AD | GoodsAD | MSD | KolektorSDD | Mastcam |
| DevNet | 0.911 | 0.723 | 0.961 | 0.916 | 0.761 |
| FastRecon | 0.941 | 0.783 | 0.954 | 0.937 | 0.802 |
| MeLIAD | **0.956** | **0.801** | **0.985** | **0.938** | **0.824** |

TABLE V: AVERAGE ROAD INTERPRETABILITY QUANTIFICATION SCORES OF IAD METHODS OVER ALL ANOMALY CATEGORIES.

| Metric | IAD Methods | | | |
|---|---|---|---|---|
| | FCDD | MKD | DevNet | MeLIAD |
| ROAD | 0.068 | 0.054 | 0.086 | **0.098** |

IAD methods, by visualizing their output heatmaps. Only IAD methods are considered for this evaluation because the heatmap generation process is an inherent part of their architecture similarly to MeLIAD. To this end, Fig. 2 displays representative anomalous images per image category, along with their respective ground truth (GT) mask. It can be observed that MeLIAD provides precise interpretation heatmaps of the detected anomalies that are more consistent with the GT masks, than those produced by the other interpretable methods. This observation is consistent with the pixel-level AUROC localization results of Table III, that are computed between the predictions and the GT masks, confirming the precision of the interpretation heatmaps generated by MeLIAD. The heatmaps of MKD and FCDD generally highlight larger regions as anomalous, that exceed the boundaries of true anomalies; DevNet offers more precise heatmaps compared to MKD and FCDD, however it still highlights parts of the image that are considered normal.

The quantitative interpretability evaluation results are reported in Table V. These indicate the average ROAD [37] score for all classes, compared among the inherently IAD methods. As it can be observed, the highest score, that indicates higher effectiveness in providing interpretations, is achieved by the proposed method of generating explanations, which further substantiates the interpretability of MeLIAD.

### D. Metric Learning Effectiveness

To demonstrate the effect of the metric learning objective to learn the embedding space as part of the proposed methodology, the *t*-Distributed Stochastic Neighbor Embedding (*t*-SNE) [39] was used, which is a well-known dimensionality reduction technique that visually represents the spread and clustering of data points in the feature representation space. In this case, the test images of the dataset are used to generate the *t*-SNE plots presented in Fig. 3, indicative of five categories. Each plot corresponds to a different image category, depicting both anomalous (purple, red) and normal (green, blue) data points, before (crosses) and after (circles) the optimization of the metric learning objective. As it can be observed, prior to the optimization process, the *t*-SNE plots, represented by the purple and green crosses showcase dispersed points in the embedding space, indicating low discriminative representations of features. After the optimization process, the *t*-SNE plots represented by the red and blue circles, depict distinct clustered groups. This suggests that the metric learning objective enhances the feature representation, by achieving better separability among normal and abnormal instances.

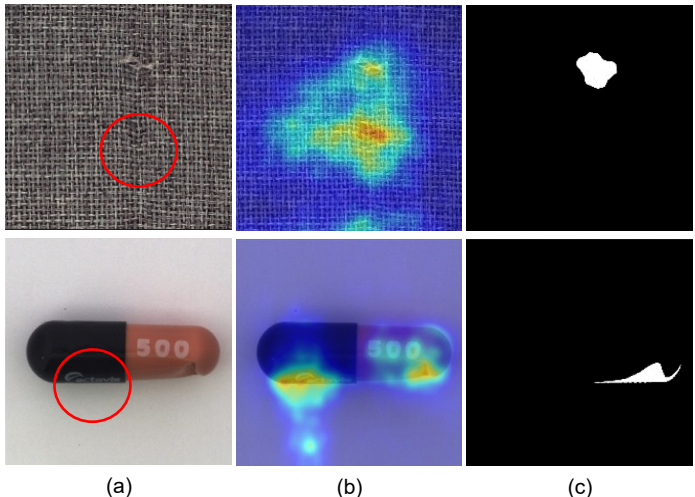

(a)         (b)         (c)

Fig. 4. Example cases of erroneous model behaviour. (a) Original images with possible anomalies (in the red circle) that have not been annotated as such in the ground truth masks, (b) Predicted interpretation heatmaps. (c) Ground truth masks.

Following the evaluation protocol of previous well-known metric learning methodologies [40], two performance metrics were computed to quantify the clustering quality of the metric learning objective. Specifically, Recall@1 signifies the percentage of data points where at least one nearest neighbor belongs to the same class. It was measured at 52.7% before and 88.2% after the optimization of the metric learning objective, indicating the capacity of the model to retrieve relevant anomalies within the top-ranked results. Also, accuracy is measured by the mean Average Precision at R (mAP@R) [41] metric, that combines the average Precision and R-precision metrics, with R denoting the number of nearest neighbors for each sample. It was measured at 43.4% before, and 75% after the optimization of metric learning, indicating the capacity of the model to identify pairs of anomalous images based on their semantic similarity.

## V. DISCUSSION

The results obtained from the experimental evaluation indicate that MeLIAD offers improved AD performance over other relevant methods on different benchmark datasets. However, this study also revealed some issues which may be considered as limitations. For example MeLIAD exhibits a sensitivity to textural variations, which could be considered normal for some image categories. This affects the overall pixel-level AUC. For example in the first row of Fig. 4, MeLIAD detects a slight variation in the carpet's texture located below the region that has been characterized as anomaly by the experts. However, there is also a possibility that this is indeed an anomaly that has been missed by the expert during the annotation process. Therefore, considering the ambiguity of whether this is an anomaly or not, such a sensitivity may not necessarily be a limitation. A similar issue appears in the classes representing objects, such as the pill image in the second row of Fig. 4. MeLIAD highlights the letters on the lower left side of the capsule as abnormal, which can be expected as these letters are absent in most of the normal images of the respective class. On the other hand, the number on the right ('500') exists in all normal images; therefore, it was not considered as an anomaly by MeLIAD.

Another issue that could be considered as a limitation, is that the number of anomalies $k$ to be included in the training set is manually determined, *e.g.*, empirically or after an ablation study, such as the one performed in section IV. Also, the selection of the most representative anomalies to be included in the training set is a challenge. However, these constitute limitations not only for MeLIAD, but also for most current few-shot visual AD methods.

## VI. Conclusions and Future Work

In this paper, MeLIAD, a novel AD methodology was presented. It requires only a few known anomalies for training and provides inherently interpretable results in the form of visual explanations, by proposing a trainable anomaly scoring entropy-based component that is jointly optimized with a metric learning objective in a unified methodology. Unlike other inherently interpretable methods MeLIAD: a) does not rely on imposing prior knowledge over the anomaly scores, that assumes the distribution of true anomalies, and b) features a trainable entropy-based anomaly scoring component. The conclusions of this study can be summarized as follows:

- The results obtained from the experimental study validate the robustness of MeLIAD, and its effectiveness, even using a very small number of images with anomalies in the training set.
- MeLIAD can achieve a higher detection and localization performance on visual quality inspection tasks, in terms of AUROC scores, when compared to state-of-the-art FSAD and IAD methodologies.
- The improved interpretability offered by MeLIAD was demonstrated both quantitatively and qualitatively, by the average ROAD scores and the interpretation heatmap comparisons, respectively.

Future work includes exploring adaptive strategies for the selection of $k$, and investigation of image mining techniques [42] for selecting representative samples to improve model performance in multimedia applications.

## References

[1] P. Wu, W. Wang, F. Chang, C. Liu, and B. Wang, "Dss-net: Dynamic self-supervised network for video anomaly detection," *IEEE Transactions on Multimedia*, 2023.

[2] V. Chandola, A. Banerjee, and V. Kumar, "Anomaly detection: A survey," *ACM computing surveys (CSUR)*, vol. 41, no. 3, pp. 1–58, 2009.

[3] C. Huang, Q. Xu, Y. Wang, Y. Wang, and Y. Zhang, "Self-supervised masking for unsupervised anomaly detection and localization," *IEEE Transactions on Multimedia*, 2022.

[4] K. Wu, L. Zhu, W. Shi, W. Wang, and J. Wu, "Self-attention memory-augmented wavelet-CNN for anomaly detection," *IEEE Transactions on Circuits and Systems for Video Technology*, vol. 33, no. 3, pp. 1374–1385, 2022.

[5] G. Pang, C. Shen, L. Cao, and A. V. D. Hengel, "Deep learning for anomaly detection: A review," *ACM computing surveys (CSUR)*, vol. 54, no. 2, pp. 1–38, 2021.

[6] S. Wei, X. Wei, Z. Ma, S. Dong, S. Zhang, and Y. Gong, "Few-shot online anomaly detection and segmentation," *Knowledge-Based Systems*, p. 112168, 2024.

[7] H. Deng, H. Luo, W. Zhai, Y. Cao, and Y. Kang, "Prioritized Local Matching Network for Cross-Category Few-Shot Anomaly Detection," *IEEE Transactions on Artificial Intelligence*, 2024.

[8] Y. Wang, Q. Yao, J. T. Kwok, and L. M. Ni, "Generalizing from a few examples: A survey on few-shot learning," *ACM computing surveys (csur)*, vol. 53, no. 3, pp. 1–34, 2020.

[9] I. Ahmed, G. Jeon, and F. Piccialli, "From artificial intelligence to explainable artificial intelligence in industry 4.0: a survey on what, how, and where," *IEEE Transactions on Industrial Informatics*, vol. 18, no. 8, pp. 5031–5042, 2022.

[10] K. Doshi and Y. Yilmaz, "Towards interpretable video anomaly detection," in *Proceedings of the IEEE/CVF Winter Conference on Applications of Computer Vision*, 2023, pp. 2655–2664.

[11] J. Feng, Y. Liang, and L. Li, "Anomaly detection in videos using two-stream autoencoder with post hoc interpretability," *Computational Intelligence and Neuroscience*, vol. 2021, 2021.

[12] C. Rudin, "Stop explaining black box machine learning models for high stakes decisions and use interpretable models instead," *Nature machine intelligence*, vol. 1, no. 5, pp. 206–215, 2019.

[13] M. Gupta, J. Gao, C. C. Aggarwal, and J. Han, "Outlier detection for temporal data: A survey," *IEEE Transactions on Knowledge and data Engineering*, vol. 26, no. 9, pp. 2250–2267, 2013.

[14] F. Ye, C. Huang, J. Cao, M. Li, Y. Zhang, and C. Lu, "Attribute restoration framework for anomaly detection," *IEEE Transactions on Multimedia*, vol. 24, pp. 116–127, 2020.

[15] R. T. Ionescu, F. S. Khan, M.-I. Georgescu, and L. Shao, "Object-centric auto-encoders and dummy anomalies for abnormal event detection in video," in *Proceedings of the IEEE/CVF Conference on Computer Vision and Pattern Recognition*, 2019, pp. 7842–7851.

[16] J. Jiang, J. Zhu, M. Bilal, Y. Cui, N. Kumar, R. Dou, F. Su, and X. Xu, "Masked swin transformer unet for industrial anomaly detection," *IEEE Transactions on Industrial Informatics*, vol. 19, no. 2, pp. 2200–2209, 2022.

[17] X. Huang, Y. Hu, X. Luo, J. Han, B. Zhang, and X. Cao, "Boosting variational inference with margin learning for few-shot scene-adaptive anomaly detection," *IEEE Transactions on Circuits and Systems for Video Technology*, 2022.

[18] Z. Wang, Y. Zhou, R. Wang, T.-Y. Lin, A. Shah, and S. N. Lim, "Few-shot fast-adaptive anomaly detection," *Advances in Neural Information Processing Systems*, vol. 35, pp. 4957–4970, 2022.

[19] S. Sheynin, S. Benaim, and L. Wolf, "A hierarchical transformation-discriminating generative model for few shot anomaly detection," in *Proceedings of the IEEE/CVF International Conference on Computer Vision*, 2021, pp. 8495–8504.

[20] Z. Fang, X. Wang, H. Li, J. Liu, Q. Hu, and J. Xiao, "FastRecon: Few-shot Industrial Anomaly Detection via Fast Feature Reconstruction," in *Proceedings of the IEEE/CVF International Conference on Computer Vision*, 2023, pp. 17481–17490.

[21] C. Huang, H. Guan, A. Jiang, Y. Zhang, M. Spratling, and Y.-F. Wang, "Registration based few-shot anomaly detection," in *European Conference on Computer Vision*, 2022, pp. 303–319.

[22] G. Pang, C. Ding, C. Shen, and A. van den Hengel, "Explainable deep few-shot anomaly detection with deviation networks," *arXiv preprint arXiv:2108.00462*, 2021.

[23] Z. Li, Y. Zhu, and M. Van Leeuwen, "A survey on explainable anomaly detection," *ACM Transactions on Knowledge Discovery from Data*, vol. 18, no. 1, pp. 1–54, 2023.

[24] J. Tritscher, A. Krause, and A. Hotho, "Feature relevance XAI in anomaly detection: Reviewing approaches and challenges," *Frontiers in Artificial Intelligence*, vol. 6, p. 1099521, 2023.

[25] S. Venkataramanan, K.-C. Peng, R. V. Singh, and A. Mahalanobis, "Attention guided anomaly localization in images," in *European Conference on Computer Vision*, 2020, pp. 485–503.

[26] G. Dimas, E. Cholopoulou, and D. K. Iakovidis, "E pluribus unum interpretable convolutional neural networks," *Scientific Reports*, vol. 13, no. 1, p. 11421, 2023.

[27] P. Liznerski, L. Ruff, R. A. Vandermeulen, B. J. Franks, M. Kloft, and K.-R. Müller, "Explainable deep one-class classification," *arXiv preprint arXiv:2007.01760*, 2020.

[28] M. Salehi, N. Sadjadi, S. Baselizadeh, M. H. Rohban, and H. R. Rabiee, "Multiresolution knowledge distillation for anomaly detection," in *Proceedings of the IEEE/CVF conference on computer vision and pattern recognition*, 2021, pp. 14902–14912.

[29] G. Pang, K. M. Ting, and D. Albrecht, "LeSiNN: Detecting anomalies by identifying least similar nearest neighbours," in *2015 IEEE international conference on data mining workshop (ICDMW)*, 2015, pp. 623–630.

[30] C.-Y. Wu, R. Manmatha, A. J. Smola, and P. Krahenbuhl, "Sampling matters in deep embedding learning," in *Proceedings of the IEEE international conference on computer vision*, 2017, pp. 2840–2848.

[31] L. Ruff, R. A. Vandermeulen, B. J. Franks, K.-R. Müller, and M. Kloft, "Rethinking assumptions in deep anomaly detection," *arXiv preprint arXiv:2006.00339*, 2020.

[32] P. Bergmann, K. Batzner, M. Fauser, D. Sattlegger, and C. Steger, "The MVTec anomaly detection dataset: a comprehensive real-world dataset for unsupervised anomaly detection," *International Journal of Computer Vision*, vol. 129, no. 4, pp. 1038–1059, 2021.

[33] J. Zhang, R. Ding, M. Ban, and L. Dai, "PKU-GoodsAD: A supermarket goods dataset for unsupervised anomaly detection and segmentation," *IEEE Robotics and Automation Letters*, 2024.

[34] J. Zhang, R. Ding, M. Ban, and T. Guo, "FDSNeT: An accurate real-time surface defect segmentation network," in *ICASSP 2022-2022 IEEE International Conference on Acoustics, Speech and Signal Processing (ICASSP)*, 2022, pp. 3803–3807.

[35] D. Tabernik, S. Šela, J. Skvarc, and D. Skocaj, "Segmentation-based deep-learning approach for surface-defect detection," *Journal of Intelligent Manufacturing*, vol. 31, no. 3, pp. 759–776, 2020.

[36] H. R. Kerner, K. L. Wagstaff, B. D. Bue, D. F. Wellington, S. Jacob, P. Horton, J. F. Bell, C. Kwan, and H. Ben Amor, "Comparison of novelty detection methods for multispectral images in rover-based planetary exploration missions," *Data Mining and Knowledge Discovery*, vol. 34, pp. 1642–1675, 2020.

[37] Y. Rong, T. Leemann, V. Borisov, G. Kasneci, and E. Kasneci, "A consistent and efficient evaluation strategy for attribution methods," *arXiv preprint arXiv:2202.00449*, 2022.

[38] K. Roth, L. Pemula, J. Zepeda, B. Schölkopf, T. Brox, and P. Gehler, "Towards total recall in industrial anomaly detection," in *Proceedings of the IEEE/CVF Conference on Computer Vision and Pattern Recognition*, 2022, pp. 14318–14328.

[39] L. Van der Maaten and G. Hinton, "Visualizing data using t-SNE.," *Journal of machine learning research*, vol. 9, no. 11, 2008.

[40] H. Oh Song, Y. Xiang, S. Jegelka, and S. Savarese, "Deep metric learning via lifted structured feature embedding," in *Proceedings of the IEEE conference on computer vision and pattern recognition*, 2016, pp. 4004–4012.

[41] K. Musgrave, S. Belongie, and S.-N. Lim, "A metric learning reality check," in *Computer Vision–ECCV 2020: 16th European Conference, Glasgow, UK, August 23–28, 2020, Proceedings, Part XXV 16*, 2020, pp. 681–699.

[42] D. K. Iakovidis, S. Tsevas, and A. Polydorou, "Reduction of capsule endoscopy reading times by unsupervised image mining," *Computerized Medical Imaging and Graphics*, vol. 34, no. 6, pp. 471–478, 2010.